\documentclass[11pt,a4paper]{article}
\usepackage[hyperref]{ranlp2023}
\usepackage{times}
\usepackage{latexsym}
\usepackage{color}
\usepackage{csquotes}
\usepackage{amsmath}
\usepackage{textcomp}
\usepackage{todonotes}
\usepackage{enumitem}
\usepackage{booktabs}
\renewcommand{\UrlFont}{\ttfamily\small}

\usepackage{microtype}

\aclfinalcopy %

\usepackage{tikz}
\usetikzlibrary{positioning}

\pgfqkeys{/mylabel}{%
    shift x/.initial = 1.5em,
    shift y/.initial = .2em,
    slope width/.initial = .3em,
    text xsep/.initial  = .1em,
    text ysep/.initial  = .1em,
    label xsep/.initial = .3333em,
    label ysep/.initial = .3333em,
    line/.initial = {thick, dashed},
    line color/.initial  = black,
    text color/.initial  = black,
    label color/.initial = black,
    color/.style = {line color=#1, label color=#1},
    /mylabel/pos/.is choice,
    /mylabel/pos/above/.style = {pos=above right},
    /mylabel/pos/below/.style = {pos=below right},
    /mylabel/pos/below right/.style = {%
        _anchor label  = north west,
        _direction     = below,
        _line pos      = south,
    },
    /mylabel/pos/above right/.style = {%
        _anchor label  = south west,
        _direction     = above,
        _line pos      = north,
    },
    _anchor label/.initial,
    _direction/.initial,
    _line pos/.initial,
    inline strut/.initial=\vphantom{Ap},
    label strut/.initial=\strut,
}

\newcommand{\setmylabel}[1]{%
    \pgfqkeys{/mylabel}{#1}%
}

\setmylabel{%
    pos = below,
}

\setmylabel{%
    line color  = blue,
    text color  = black,
    label color = purple,
}

\usepackage{xspace}
\usepackage{booktabs}
\usepackage{graphicx}
\usepackage{enumitem}
\usepackage{xcolor}
\usepackage{tikz-dependency}
\usepackage{url}
\usepackage{amssymb}
\usepackage{float}
\usepackage{todonotes}
\usepackage{csquotes}
\usepackage{xspace}
\usepackage{listings}
\usepackage{caption}
\usepackage{subcaption}
\usepackage{soul} %
\usepackage[normalem]{ulem}

\setlist[description]{leftmargin=\parindent,labelindent=0pt}

\usepackage{tablefootnote}
\usepackage{tabularx}
\newcolumntype{L}{>{\raggedright\arraybackslash}X}
\usepackage{amsmath}
\usepackage{tikz}

\newcommand{\tref}[1]{Table~\ref{#1}}
\newcommand{\fref}[1]{Figure~\ref{#1}}

\newcommand{\roberta}{RoBERTa}

\makeatletter
\def\url@leostyle{%
  \@ifundefined{selectfont}{\def\UrlFont{\sf}}{\def\UrlFont{\scriptsize\sffamily}}}
\makeatother
\newcommand{\CNC}{Causal News Corpus}

\newcommand{\argzero}{\texttt{\textlangle ARG0\textrangle}}
\newcommand{\argone}{\texttt{\textlangle ARG1\textrangle}}
\newcommand{\sig}{\texttt{\textlangle SIG0\textrangle}}

\newcommand{\Stone}{Subtask 1}

\newcommand{\stone}{Subtask~1\xspace} %
\newcommand{\sttwo}{Subtask~2\xspace}

\title{BoschAI @ Causal News Corpus 2023: Robust Cause-Effect Span Extraction using Multi-Layer Sequence Tagging and Data Augmentation}

\author{Timo Pierre Schrader$^{1,2}$~
  Simon Razniewski$^1$~
  Lukas Lange$^{1}$~
  Annemarie Friedrich$^2$\\
  $^1$Bosch Center for Artificial Intelligence, Renningen, Germany \\ 
    $^2$University of Augsburg, Germany \\
\texttt{timo.schrader|simon.razniewski|lukas.lange@de.bosch.com} \\
  \texttt{annemarie.friedrich@informatik.uni-augsburg.de}}

\date{}

\begin{document}
\maketitle

\begin{abstract}
Understanding causality is a core aspect of intelligence.
The Event Causality Identification with Causal News Corpus Shared Task addresses two aspects of this challenge: \stone aims at detecting causal relationships in texts, and \sttwo requires identifying signal words and the spans that refer to the cause or effect, respectively.
Our system, which is based on pre-trained transformers, stacked sequence tagging, and synthetic data augmentation, ranks third in \stone and wins \sttwo with an F1 score of 72.8, corresponding to a margin of 13 pp. to the second-best system. %

\end{abstract}

\section{Introduction}
\label{sec:intro}

In this paper, we describe our approach to the Event Causality Identification with Causal News Corpus shared task \citep{tan-2023-event}, which took place at The 6th Workshop on Challenges and Applications of Automated Extraction of Socio-political Events from Text (CASE 2023).
The task, which builds on the 2022 iteration of the same shared task \citep{tan-etal-2022-event}, but including more labeled data, targets the detection and extraction of causal relationships.  %
In \stone, participating systems need to decide whether a sentence contains any causal relationship.
\sttwo requires extracting the spans that denote cause, effect, and trigger words (if any).

Our system leverages pre-trained transformer encoders and synthetic data augmentation methods, and ranks third in \stone.
We address \sttwo using a supervised sequence labeling model, which wins by a margin of 13 percentage points in terms of F1 over the second-best system.
We model multiple causal chains per sentence via stacked labels and find that synthetic data augmentation consistently improves performance.
Our code is publicly available.\footnote{\url{https://github.com/boschresearch/boschai-cnc-shared-task-ranlp2023}}

\begin{figure}[t]
    \centering
    \includegraphics[width=.85\linewidth]{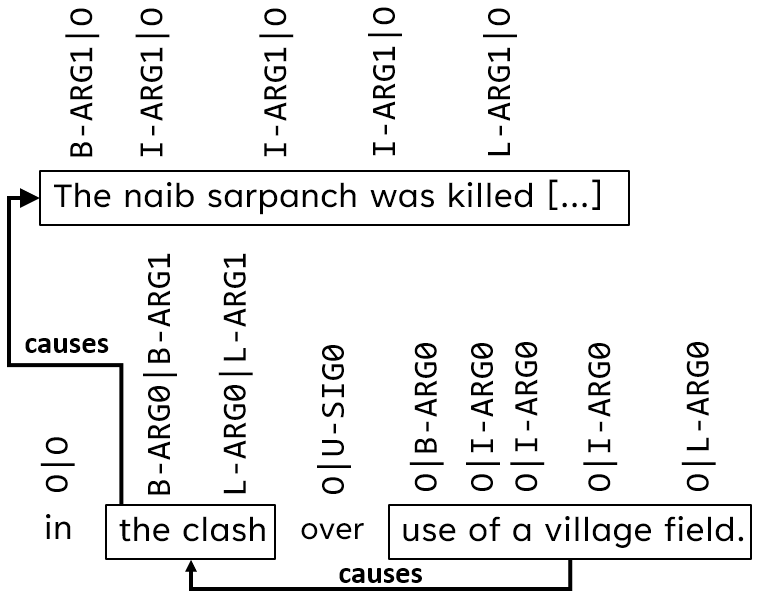}
    \caption{Our proposed modeling technique for extracting causal relationships (\sttwo) using stacked BILOU labels. \texttt{ARG0} = cause, \texttt{ARG1} = effect.}
    \label{fig:teaser}
\end{figure}
\section{Dataset and Task}
\label{sec:dataset}
The \CNC~
\citep[CNC,][]{tan-etal-2022-causal}
consists of 3767 sentences extracted from news articles.
CNC provides annotations of semantic relations of the form \enquote{\textit{X} causes \textit{Y}} that indicate a \textit{causal} relationship between arguments \textit{X} and \textit{Y}.
The definition of causality follows that of the \textsc{Contingency} label in the PDTB-3 corpus \citep{webber2019penn}, which is used when a statement provides the reason, explanation, or justification for another event.
Following TimeML \citep{pustejovsky-et-al}, %
the definition of events includes both actions that happen or occur and states.
As illustrated by the example in \fref{fig:teaser}, one event is the immediate effect of another, e.g., the event expressed by \enquote{the use of a village field} is the cause of that expressed by \enquote{the clash.}

While following the definition of causal relations of PDTB-3, which focuses on causal relations between sentences or clauses, CNC provides span annotations for causes (\texttt{ARG0}), effects (\texttt{ARG1}) and signals (\texttt{SIG0}) within sentences.
Spans may comprise one to several words. Their boundaries are not restricted to clause or constituent boundaries.
Signals are expressions such as \enquote{has led to} or \enquote{causing,} but not every causal relation annotation requires a signal.
Of all annotated relations, 30\% do not contain a signal, for example: \enquote{[Dissatisfied with the package\textsubscript{Cause}], [workers staged an all-night sit-in\textsubscript{Effect}].}
The average signal length is $1.46$ words.
\citet{tan-etal-2022-causal} describe the annotation guidelines in detail.

The shared task is divided into two subtasks:
Subtask 1 is a binary classification problem, deciding whether a sentence contains a cause-effect chain or not.
Subtask 2 deals with the more challenging problem of extracting the correct spans of cause, effect, and signal, where a sentence may contain more than one causal relation.
In CNC, the maximum number of causal relations per sentence is four.
Spans are annotated using XML-like tags: \argzero~refers to causes, \argone~to effects, and \sig~to signals.

\section{Modeling and Augmentation}
\label{sec:comp_modeling}
In this section, we describe the neural architectures that we use to solve the two subtasks.
To produce contextualized embeddings of the input sentences, we use BERT-Large \citep{devlin-etal-2019-bert} and RoBERTa-Large \citep{liu2019roberta}.

\subsection{Subtask 1}
We implement a binary classifier to detect whether a sentence contains a cause-effect relation.
The sentence-level \texttt{[CLS]} embedding is fed into a linear output layer that outputs a prediction on whether a sentence contains a cause-effect meaning or not.
We design the output layer to yield two prediction scores, one for each class.
During our experiments, we observe that the classifier has shown prediction bias towards negative samples.
Hence, we apply a weighted cross entropy loss that upweights the positive samples.%

\subsection{Subtask 2}
We model the problem of detecting cause, effect, and signal spans, potentially with multiple causal relations within a single sentence, as sequence tagging task using the BILOU labeling scheme \citep{alex-etal-2007-recognising}.
The BILOU scheme extends the commonly used BIO scheme by introducing two additional markers, where \enquote{L} denotes the end of a multi-token sequence and \enquote{U} refers to a single-token entity.
For example, the argument span \enquote{Beijing launched a campaign} has the label sequence \texttt{[B-ARG1 I-ARG1 I-ARG1 L-ARG1]} (ignoring BERT-specific subword tokens here).
A linear layer on top of the embedding model produces the logits for all BILOU tags for each token individually.
These logits are fed into a conditional random field \citep[CRF,][]{lafferty2001CRF} output layer, which computes the most likely consistent tag sequence.%

However, this approach can only predict a single output sequence per sample, i.e., is not able to detect multiple causal chains in an instance.
Consider the example shown in \fref{fig:teaser}. %
The expression \enquote{the clash} can be either the cause of one killing and 17 injuries or the effect of not being able to agree about the usage of a village field.
As a result, there are two causal relations within this instance.
To address this, we \enquote{stack} the BILOU labels by concatenating them using a pipe (\enquote{\texttt{|}}) operator, similar to \citet{strakova-etal-2019-neural}, who also use a label stacking approach.
As shown in \fref{fig:teaser}, this means that the word \enquote{clash} is tagged with \texttt{L-ARG0|L-ARG1|O}, which decodes to being the end of a cause in the first layer, being the end of an effect in the second one and not being part of any span in the third one.

To keep the label space manageable, we model three layers.
There are only nine samples in the training set with four possible sequences. %
Without filtering, we would end up with about 39,000 labels.
We only add stacked labels that occur in the training and validation data, resulting in roughly 300 three-layer BILOU labels.
During evaluation, these stacked labels are split into their three distinct layers and each instance is evaluated separately.
As a result, the model is able to predict up to three different causal relations per sentence.

\begin{table*}[!t]
   \centering
    \setlength{\tabcolsep}{0.8em}\renewcommand{\arraystretch}{1.7}%
    \newcolumntype{L}{>{\raggedright\arraybackslash}X}
    \small
    \begin{tabularx}{\textwidth}{L|X}
    \toprule
    \textbf{Original Sentence} & \textbf{EDA Augmented Sentence} \\
    \midrule
    His arrest has sparked widespread \ul{protests by students, teachers as well as opposition} parties.
    & His arrest has sparked widespread \ul{resist by student, teacher as advantageously as confrontation} parties. \\
    Month-long \ul{escalating protests to mark 4th anniversary} of Mullivaikkal pogrom. & Month-long \ul{step up protests to mark off quaternary day of remembrance} of Mullivaikkal pogrom. \\
    They \ul{also rubbished} suggestions that the \ul{student} protests were \ul{losing} steam [...] & They \ul{besides rubbish} suggestions that the \ul{scholar} protests were \ul{lose} steam [...] \\
    \bottomrule
    \end{tabularx}
    \caption{Comparison between original sentences and their EDA-augmented counterparts. Differences are underlined.}
    \label{tab:augmented_samples}
\end{table*}

\subsection{Data Augmentation and Resampling}
As for both subtasks, there is only limited training data available, we incorporate additional synthetic data into the training. %
In the 2022 edition of the shared task, several teams also experimented with data augmentation methods.
\citet{chen-etal-2022-1cademy} trained BART \citep{lewis-etal-2020-bart} to rephrase instances in the dataset.
\citet{kim-etal-2022-snu} create additional data by adding the SemEval-2010 dataset \citep{hendrickx-etal-2010-semeval} and replacing words by their POS tag. %

\paragraph{Augmenting using EDA}
Our first augmentation approach makes use of the Easy Data Augmentation \citep[EDA,][]{wei-zou-2019-eda} tool to generate additional training data for both subtasks.
EDA offers different augmentation techniques: synonym replacement (\textit{sr}), random word insertion (\textit{ri}), random word deletion (\textit{rd}), and random word swaps (\textit{rs}).
The percentage of words on which these techniques are applied are defined by hyperparameters $\alpha_{sr}$, $\alpha_{ri}$, $\alpha_{rd}$, and $\alpha_{rs}$.

For \stone, we employ synonym replacement, random word insertion, and random swaps and generate four synthetic samples per original instance in the training set. This results in a training set five times as large as the original dataset with a total sample count of over 15.000 samples.\footnote{We noticed that the tool also clones each original sample in our implementation.}
In \sttwo, keeping the ordering of \argzero, \argone~and \sig~consistent is of high importance.
To avoid adding destructive noise to the training data, we only use synonym replacement and random insertion for this subtask.
We add one augmented sample per single-relation instance, i.e., we do not augment data based on samples with more than one causal relation.
We discard augmented samples that are invalid w.r.t. the annotation scheme.
Data augmentation for the challenging multi-relation cases is an interesting direction for future research.
The augmented training set contains 4.611 instances, i.e., about 1.500 more than the original set.

\tref{tab:augmented_samples} shows three instances and their augmented counterparts.
The first example shows a replacement of \textit{opposition} by \textit{confrontation}, which is not fully synonymous, but still related.
In the second one, there is a synonym replacement of \textit{4th} by \textit{quaternary}.
In the third example, noise is added by replacing \enquote{losing} with \enquote{lose}, illustrating that the data augmentation method does not control for grammatical correctness.

\paragraph{Oversampling of Multi-Relation Samples}
About 32\% of all instances with at least one causal relation in the training set are labeled with more than one causal relation.
Out of these, we sample 400 instances (with replacement) and add them to the training dataset.
In contrast to EDA, we only use this setting only for \sttwo.

\paragraph{Generating Samples using ChatGPT}
\label{sssec:chatgpt}
We experiment with GPT-3.5-turbo %
and prompt it to generate 100 novel samples containing causal relations that are similar to those of the CNC corpus.
We prompt ChatGPT with multiple samples of the CNC train set, and the rules of placing \argzero, \argone, and \sig, and let it generate novel samples.
This additional data is only used for \sttwo.

The ChatGPT-based data augmentation approach generates relatively simple examples by always sticking to a Cause-Signal-Effect or Effect-Signal-Cause structure without overlapping spans.
Examples include %
\enquote{[The lack of rain\textsubscript{Cause}] [caused\textsubscript{Signal}] [the crops to fail and farmers to suffer losses\textsubscript{Effect}].}
and \enquote{[A decrease in greenhouse gas emissions\textsubscript{Effect}] [was a result of\textsubscript{Signal}] [the decrease in demand for fossil fuels\textsubscript{Cause}]}.

\section{Experimental Evaluation}
\label{sec:results}

This section describes our experimental results for both subtasks.
Evaluation of \sttwo~is performed using FairEval\footnote{\url{https://huggingface.co/spaces/hpi-dhc/FairEval/tree/main}}, which implements a relaxation of traditional hard-matching span evaluation metrics on sentences marked as containing a causal relation in the gold standard only.
We train our on all samples of the train split, including those without causal relations.
\subsection{Hyperparameters}
\label{ssec:hyperparams}
To find the best learning rates and augmentation parameter %
combinations, we employ a grid search and refine the learning rates after an initial coarse-grained search ranging from $1e^{-7}$ to $9e^{-4}$ for the pre-trained language model. %
The binary classifier for \stone is trained with a learning rate of 8e-6, using the EDA augmented training data and a batch size of 32.
For \stone, we use the following parameter values for the different EDA techniques: $\alpha_{sr} = 0.4$, $\alpha_{ri} = 0.1$, and $\alpha_{rs} = 0.6$.
We use a weighted cross entropy loss for this subtask, using a weight of 1.5 for %
class \textit{causal}. %
For \sttwo, we apply the following settings: $\alpha_{sr} = 0.4$ and $\alpha_{ri} = 0.5$.

The CRF-based tagger for \sttwo uses a learning rate of $7e^{-5}$ for the language model and the linear layer, whereas a learning rate of $3e^{-4}$ is applied on the CRF.
During fine-tuning, EDA-augmentated data is included in the training set.
Training the models is performed on Nvidia A100 GPUs using one GPU per run, which takes several hours per model.
Early stopping is applied using the F1 score on the dev set and a patience of three epochs to select the best model.
The models are optimized using AdamW \citep{loshchilov2019decoupled} and an inverse square-root learning rate scheduler taken from \citet{grunewald-etal-2021-applying}.

\begin{table}[t]
\centering
\footnotesize
\begin{tabular}{llccc}
\toprule
 & \textbf{Team}& \textbf{Precision} & \textbf{Recall} & \textbf{F1} \\
 \midrule
1 & DeepBlueAI & \textbf{83.2} & 86.1 & \textbf{84.7} \\
2 & InterosML & 81.6 & 87.3 & 84.4 \\
3 & BoschAI & 80.0 & 87.9 & 83.8 \\
\cmidrule(lr){2-5}
& \textit{baseline} & 75.9 & \textbf{89.2} & 81.9 \\
\bottomrule
\end{tabular}
\caption{\Stone: results on \textbf{test} of the best three systems and the baseline provided by \citet{tan-2023-event}. Scores are based on the public leaderboard.}
\label{tab:st1_results}
\end{table}

\begin{table}[t]
\centering
\footnotesize
\begin{tabular}{lcccc}
\toprule
\textbf{LM}& \textbf{Precision} & \textbf{Recall} & \textbf{F1} & \textbf{Accuracy} \\
 \midrule
BERT & 86.9 & \textbf{89.7} & \textbf{88.3} & 87.1 \\
RoBERTa & \textbf{88.6} & 88.1 & \textbf{88.3} & \textbf{87.4} \\
\bottomrule
\end{tabular}
\caption{\Stone: results on \textbf{dev} (large model variants).}
\label{tab:st1_dev_results_bert_roberta}
\end{table}

\subsection{Results}

\begin{table}[!tp]
\centering
\footnotesize
\setlength\tabcolsep{4pt}
\begin{tabular}{llccc|ccc}
\toprule
& & \multicolumn{3}{c}{\textbf{All relations}} & \multicolumn{3}{c}{\textbf{Multi-relation}}\\ %
& & \textbf{P} & \textbf{R} & \textbf{F1} & \textbf{P} & \textbf{R} & \textbf{F1} \\
\midrule
1 & BoschAI & \textbf{84.4} & \textbf{64.0} & \textbf{72.8} & 82.6 & 53.5 & 64.9\\
\cmidrule(lr){2-8}
& - Cause & 85.3 & 59.7 & 70.2 & 82.5 & 47.4 & 60.2 \\
& - Effect & 82.8 & 62.9 & 71.5 & 80.3 & 50.4 & 61.9 \\
& - Signal & 85.4 & 70.4 & 77.2 & 82.6 & 53.5 & 64.9\\
\midrule
2 & tanfiona* & 60.3 & 59.2 & 59.7 & - & - & - \\
3 & CSECU-DSG & 40.0 & 36.1 & 38.0 & - & - & - \\
\bottomrule
\end{tabular}
\caption{Per-class scores on the \textbf{test} for \sttwo~ of our best scoring model using \roberta-Large and EDA. The last two rows show the results of the second- and third-best system. *System of \citet{chen-etal-2022-1cademy}.}
\label{tab:st2_results}
\end{table}

In the following, we refer to the public leaderboard of the Event Causality Identification with Causal News Corpus shared task.\footnote{\url{https://codalab.lisn.upsaclay.fr/competitions/11784\#results}}
We report results on test as provided by the leaderboard evaluation script.

\paragraph{Subtask 1}

Our \roberta-based binary classifier ranks third of 10 participants.
Results are shown in \tref{tab:st1_results}, including the best two systems and the baseline by \citet{tan-2023-event}.
Among the top three, we achieve the best recall score. %
Qualitatively, we find that neither sentence length nor the presence of signal words are strongly correlated with misclassifications. %

We report the results of our classifier that uses BERT-Large in comparison to \roberta-Large in \tref{tab:st1_dev_results_bert_roberta} on the dev set (since we do not have access to the gold standard of test).
Both models perform almost equally on this task, with \roberta~outperforming BERT by a slight margin in terms of accuracy with a difference 0.3\% pp.

\begin{table*}[!tp]
    \centering
    \footnotesize
    \setlength\tabcolsep{7pt}
    \begin{tabular}{l|ccc|ccc|ccc|ccc}
        \toprule
        {} & \multicolumn{3}{c}{\textbf{Cause}} & \multicolumn{3}{c}{\textbf{Effect}} & \multicolumn{3}{c}{\textbf{Signal}} & \multicolumn{3}{c}{\textit{avg}}\\
         \textbf{LM} & \textbf{P} & \textbf{R} & \textbf{F1} & \textbf{P} & \textbf{R} & \textbf{F1} & \textbf{P} & \textbf{R} & \textbf{F1} & \textbf{P} & \textbf{R} & \textbf{F1} \\
         \midrule
         BERT-Large & 82.4 & 59.9 & 69.4 & 83.2 & 58.7 & 68.9 & 86.3 & 72.0 & 78.5 & 83.8 & 62.6 & 71.6 \\
         \midrule
         RoBERTa-Large & 86.8 & 66.1 & 75.1 & 85.2 & \textbf{68.5} & 76.0 & 82.8 & 75.4 & 78.9 & 85.1 & 69.3 & 76.4 \\ %
         + EDA* & 86.4 & \textbf{67.9} & 76.1 & \textbf{88.5} & 67.9 & \textbf{76.8} & 85.0 & \textbf{77.5} & \textbf{81.1} & 86.8 & \textbf{70.3} & \textbf{77.7} \\ %
         + Oversampling & 87.5 & 67.7 & \textbf{76.3} & 86.8 & 66.2 & 75.1 & \textbf{85.1} & 74.3 & 79.3 & 86.6 & 68.8 & 76.7\\
         + ChatGPT & \textbf{88.4} & 65.7 & 75.4 & 87.5 & 66.8 & 75.8 & 84.3 & 75.2 & 79.5 & \textbf{86.9} & 68.5 & 76.6\\
         \bottomrule
    \end{tabular}
    \caption{Subtask 2 results on \textbf{dev}: precision, recall and F1 scores for cause, effect and signal span predictions. *Our system used to produce leaderboard scores.}
    \label{tab:st2_all_results}
\end{table*}

\paragraph{Subtask 2}

On this task, we compare our models against the baseline provided by \citet{tan-2023-event}, which is the best performing system from the previous iteration of the shared task by team \enquote{1Cademy} \citep{chen-etal-2022-1cademy}.
They also build upon a BERT-based embedding model, but output prediction scores for begin and end tokens of the respective spans.
In order to produce consistent output, i.e., non-overlapping cause and effect spans and correctly ordered spans, they implement a beam-search algorithm on top that aims to find the top $m$ most likely spans for each of the three types.

Per-label scores of our best-performing model and those of the other two competitors are shown in \tref{tab:st2_results}.
Our best system is based on \roberta-Large with a CRF layer on top and trained on EDA-augmented data.
Our system clearly outperforms the last year's winning system by more than 13 percentage points in terms of F1 on the latest CNC data, exceeding precision by 24 percentage points. %
Our system performs best on the signal label, which could be explained by two factors: signals are much more repetitive in the corpus (with \enquote{to} occurring 293 times in the train data) and the average length of 1.46 words is much smaller than those of causes (11.74) and effects (10.74).
\tref{tab:st2_results} also lists the results for multi-relation instances only, showing that recall drops for those instances. %

\tref{tab:st2_all_results} compares several settings, including various data augmentation techniques, by label on the dev set. %
We evaluate on the dev set because we do not have access to the gold standard of the test set.%

First of all, using \roberta~over BERT improves the average F1 score by 4.8 points in terms of F1. %
Next, all three data augmentation methods contribute performance improvements over the \roberta~baseline with the recall of \textbf{Effect} being the only exception.
Best overall results are achieved using EDA augmentation. %
However, ChatGPT-augmented significantly improves precision of \textbf{Cause} (1.6 points F1 over baseline) and also yields the best average precision.
\citet{tan-etal-2022-causal} also experiment with using two additional corpora, however, they do not get significant improvements, likely due to more different foci of the datasets.
The synthetic data augmentation methods that we used have the advantage of producing training data very similar to CNC.

\begin{table}[t]
\centering
\footnotesize
\begin{tabular}{cccc}
\toprule
 \textbf{Relations/Sentence}& \textbf{Cause} & \textbf{Effect} & \textbf{Signal} \\
 \midrule
1 & 85.5 & 80.9 & 84.7 \\
2 & 67.7 & 76.1 & 84.0 \\
3 & 52.8 & 61.8 & 57.9 \\
\bottomrule
\end{tabular}
\caption{Per-class F1 scores by the numbers of causal relations per sentence on \textbf{dev} for \sttwo.}
\label{tab:st2_split_results}
\end{table}

Finally, \tref{tab:st2_split_results} breaks down results on dev split by single-relation, two-relation and three-relation instances.
While scores for Effect and Signal remain high for two-relation instances, performance is much smaller (yet still strong) for three-relation instances.

\section{Conclusion and Outlook}
\label{sec:conclusion}
In this paper, we have described our modeling approach to the \enquote{Event Causality Identification with Causal News Corpus} shared task (CASE 2023).
We have proposed a multi-layer sequence tagging model that aims at identifying causal relations within news-related sentences.
Our approach significantly outperforms all participating systems in \sttwo. %
Furthermore, we have shown that synthetic data augmentation methods are beneficial for this task. %
Our results indicate that  %
careful modeling, more advanced data augmentation, and leveraging larger language models may be fruitful directions for further improvements.

\bibliographystyle{acl_natbib}
\bibliography{anthology,ranlp2023}

\begin{thebibliography}{17}
\expandafter\ifx\csname natexlab\endcsname\relax\def\natexlab#1{#1}\fi

\bibitem[{Alex et~al.(2007)Alex, Haddow, and
  Grover}]{alex-etal-2007-recognising}
Beatrice Alex, Barry Haddow, and Claire Grover. 2007.
\newblock \href {https://aclanthology.org/W07-1009} {Recognising nested named
  entities in biomedical text}.
\newblock In \emph{Biological, translational, and clinical language
  processing}, pages 65--72, Prague, Czech Republic. Association for
  Computational Linguistics.

\bibitem[{Chen et~al.(2022)Chen, Zhang, Nik, Li, and
  Fu}]{chen-etal-2022-1cademy}
Xingran Chen, Ge~Zhang, Adam Nik, Mingyu Li, and Jie Fu. 2022.
\newblock \href {https://aclanthology.org/2022.case-1.14} {1{C}ademy @ causal
  news corpus 2022: Enhance causal span detection via beam-search-based
  position selector}.
\newblock In \emph{Proceedings of the 5th Workshop on Challenges and
  Applications of Automated Extraction of Socio-political Events from Text
  (CASE)}, pages 100--105, Abu Dhabi, United Arab Emirates (Hybrid).
  Association for Computational Linguistics.

\bibitem[{Devlin et~al.(2019)Devlin, Chang, Lee, and
  Toutanova}]{devlin-etal-2019-bert}
Jacob Devlin, Ming-Wei Chang, Kenton Lee, and Kristina Toutanova. 2019.
\newblock \href {https://doi.org/10.18653/v1/N19-1423} {{BERT}: Pre-training of
  deep bidirectional transformers for language understanding}.
\newblock In \emph{Proceedings of the 2019 Conference of the North {A}merican
  Chapter of the Association for Computational Linguistics: Human Language
  Technologies, Volume 1 (Long and Short Papers)}, pages 4171--4186,
  Minneapolis, Minnesota. Association for Computational Linguistics.

\bibitem[{Gr{\"u}newald et~al.(2021)Gr{\"u}newald, Friedrich, and
  Kuhn}]{grunewald-etal-2021-applying}
Stefan Gr{\"u}newald, Annemarie Friedrich, and Jonas Kuhn. 2021.
\newblock \href {https://doi.org/10.18653/v1/2021.iwpt-1.13} {Applying
  occam{'}s razor to transformer-based dependency parsing: What works, what
  doesn{'}t, and what is really necessary}.
\newblock In \emph{Proceedings of the 17th International Conference on Parsing
  Technologies and the IWPT 2021 Shared Task on Parsing into Enhanced Universal
  Dependencies (IWPT 2021)}, pages 131--144, Online. Association for
  Computational Linguistics.

\bibitem[{Hendrickx et~al.(2010)Hendrickx, Kim, Kozareva, Nakov,
  {\'O}~S{\'e}aghdha, Pad{\'o}, Pennacchiotti, Romano, and
  Szpakowicz}]{hendrickx-etal-2010-semeval}
Iris Hendrickx, Su~Nam Kim, Zornitsa Kozareva, Preslav Nakov, Diarmuid
  {\'O}~S{\'e}aghdha, Sebastian Pad{\'o}, Marco Pennacchiotti, Lorenza Romano,
  and Stan Szpakowicz. 2010.
\newblock \href {https://aclanthology.org/S10-1006} {{S}em{E}val-2010 task 8:
  Multi-way classification of semantic relations between pairs of nominals}.
\newblock In \emph{Proceedings of the 5th International Workshop on Semantic
  Evaluation}, pages 33--38, Uppsala, Sweden. Association for Computational
  Linguistics.

\bibitem[{Kim et~al.(2022)Kim, Choe, and Lee}]{kim-etal-2022-snu}
Juhyeon Kim, Yesong Choe, and Sanghack Lee. 2022.
\newblock \href {https://aclanthology.org/2022.case-1.6} {{SNU}-causality lab @
  causal news corpus 2022: Detecting causality by data augmentation via
  part-of-speech tagging}.
\newblock In \emph{Proceedings of the 5th Workshop on Challenges and
  Applications of Automated Extraction of Socio-political Events from Text
  (CASE)}, pages 44--49, Abu Dhabi, United Arab Emirates (Hybrid). Association
  for Computational Linguistics.

\bibitem[{Lafferty et~al.(2001)Lafferty, McCallum, and
  Pereira}]{lafferty2001CRF}
John~D. Lafferty, Andrew McCallum, and Fernando C.~N. Pereira. 2001.
\newblock \href {https://doi.org/10.5555/645530.655813} {Conditional random
  fields: Probabilistic models for segmenting and labeling sequence data}.
\newblock In \emph{Proceedings of the Eighteenth International Conference on
  Machine Learning {(ICML} 2001), Williams College, Williamstown, MA, USA, June
  28 - July 1, 2001}, pages 282--289. Morgan Kaufmann.

\bibitem[{Lewis et~al.(2020)Lewis, Liu, Goyal, Ghazvininejad, Mohamed, Levy,
  Stoyanov, and Zettlemoyer}]{lewis-etal-2020-bart}
Mike Lewis, Yinhan Liu, Naman Goyal, Marjan Ghazvininejad, Abdelrahman Mohamed,
  Omer Levy, Veselin Stoyanov, and Luke Zettlemoyer. 2020.
\newblock \href {https://doi.org/10.18653/v1/2020.acl-main.703} {{BART}:
  Denoising sequence-to-sequence pre-training for natural language generation,
  translation, and comprehension}.
\newblock In \emph{Proceedings of the 58th Annual Meeting of the Association
  for Computational Linguistics}, pages 7871--7880, Online. Association for
  Computational Linguistics.

\bibitem[{Liu et~al.(2019)Liu, Ott, Goyal, Du, Joshi, Chen, Levy, Lewis,
  Zettlemoyer, and Stoyanov}]{liu2019roberta}
Yinhan Liu, Myle Ott, Naman Goyal, Jingfei Du, Mandar Joshi, Danqi Chen, Omer
  Levy, Mike Lewis, Luke Zettlemoyer, and Veselin Stoyanov. 2019.
\newblock \href {http://arxiv.org/abs/1907.11692} {Roberta: A robustly
  optimized bert pretraining approach}.

\bibitem[{Loshchilov and Hutter(2019)}]{loshchilov2019decoupled}
Ilya Loshchilov and Frank Hutter. 2019.
\newblock \href {http://arxiv.org/abs/1711.05101} {Decoupled weight decay
  regularization}.

\bibitem[{Pustejovsky et~al.(2003)Pustejovsky, Castaño, Ingria, Saurí,
  Gaizauskas, Setzer, and Katz}]{pustejovsky-et-al}
James Pustejovsky, José Castaño, Robert Ingria, Roser Saurí, Rob Gaizauskas,
  Andrea Setzer, and Graham Katz. 2003.
\newblock Timeml: A specification language for temporal and event expressions.

\bibitem[{Strakov{\'a} et~al.(2019)Strakov{\'a}, Straka, and
  Hajic}]{strakova-etal-2019-neural}
Jana Strakov{\'a}, Milan Straka, and Jan Hajic. 2019.
\newblock \href {https://doi.org/10.18653/v1/P19-1527} {Neural architectures
  for nested {NER} through linearization}.
\newblock In \emph{Proceedings of the 57th Annual Meeting of the Association
  for Computational Linguistics}, pages 5326--5331, Florence, Italy.
  Association for Computational Linguistics.

\bibitem[{Tan et~al.(2022{\natexlab{a}})Tan, Hettiarachchi,
  H{\"u}rriyeto{\u{g}}lu, Caselli, Uca, Liza, and
  Oostdijk}]{tan-etal-2022-event}
Fiona~Anting Tan, Hansi Hettiarachchi, Ali H{\"u}rriyeto{\u{g}}lu, Tommaso
  Caselli, Onur Uca, Farhana~Ferdousi Liza, and Nelleke Oostdijk.
  2022{\natexlab{a}}.
\newblock \href {https://aclanthology.org/2022.case-1.28} {Event causality
  identification with causal news corpus - shared task 3, {CASE} 2022}.
\newblock In \emph{Proceedings of the 5th Workshop on Challenges and
  Applications of Automated Extraction of Socio-political Events from Text
  (CASE)}, pages 195--208, Abu Dhabi, United Arab Emirates (Hybrid).
  Association for Computational Linguistics.

\bibitem[{Tan et~al.(2023)Tan, Hettiarachchi, H{\"u}rriyeto{\u{g}}lu, Uca,
  Liza, and Oostdijk}]{tan-2023-event}
Fiona~Anting Tan, Hansi Hettiarachchi, Ali H{\"u}rriyeto{\u{g}}lu, Onur Uca,
  Farhana~Ferdousi Liza, and Nelleke Oostdijk. 2023.
\newblock Event causality identification with causal news corpus - shared task
  3, {CASE} 2023.
\newblock In \emph{Proceedings of the 6th Workshop on Challenges and
  Applications of Automated Extraction of Socio-political Events from Text
  (CASE)}. Association for Computational Linguistics.

\bibitem[{Tan et~al.(2022{\natexlab{b}})Tan, H{\"u}rriyeto{\u{g}}lu, Caselli,
  Oostdijk, Nomoto, Hettiarachchi, Ameer, Uca, Liza, and
  Hu}]{tan-etal-2022-causal}
Fiona~Anting Tan, Ali H{\"u}rriyeto{\u{g}}lu, Tommaso Caselli, Nelleke
  Oostdijk, Tadashi Nomoto, Hansi Hettiarachchi, Iqra Ameer, Onur Uca,
  Farhana~Ferdousi Liza, and Tiancheng Hu. 2022{\natexlab{b}}.
\newblock \href {https://aclanthology.org/2022.lrec-1.246} {The causal news
  corpus: Annotating causal relations in event sentences from news}.
\newblock In \emph{Proceedings of the Thirteenth Language Resources and
  Evaluation Conference}, pages 2298--2310, Marseille, France. European
  Language Resources Association.

\bibitem[{Webber et~al.(2019)Webber, Prasad, Lee, and Joshi}]{webber2019penn}
Bonnie Webber, Rashmi Prasad, Alan Lee, and Aravind Joshi. 2019.
\newblock The penn discourse treebank 3.0 annotation manual.
\newblock \emph{Philadelphia, University of Pennsylvania}, 35:108.

\bibitem[{Wei and Zou(2019)}]{wei-zou-2019-eda}
Jason Wei and Kai Zou. 2019.
\newblock \href {https://doi.org/10.18653/v1/D19-1670} {{EDA}: Easy data
  augmentation techniques for boosting performance on text classification
  tasks}.
\newblock In \emph{Proceedings of the 2019 Conference on Empirical Methods in
  Natural Language Processing and the 9th International Joint Conference on
  Natural Language Processing (EMNLP-IJCNLP)}, pages 6382--6388, Hong Kong,
  China. Association for Computational Linguistics.

\end{thebibliography}

\end{document}